\documentclass[
]{ceurart}

\usepackage{listings}
\usepackage{inputenc}
\begin{document}

\copyrightyear{2022}
\copyrightclause{Copyright for this paper by its authors.
  Use permitted under Creative Commons License Attribution 4.0
  International (CC BY 4.0).}

\conference{MediaEval'22: Multimedia Evaluation Workshop,
  January 13--15, 2023, Bergen, Norwa,y and Online}

\title{Relevance Classification of Flood-related Twitter Posts via Multiple Transformers}


\author[1]{Wisal Mukhtiar}
    \fnmark[1]

\author[1]{Waliiya Rizwan}
\fnmark[1]

\author[1]{Aneela Habib}
\fnmark[1]

\author[1]{Yasir Saleem Afridi}
\author[1]{Laiq Hasan}
\author[2]{Kashif Ahmad} [%
    email=kashif.ahmad@mtu.ie
]

\address[1]{Department of Computer Systems Engineering, University of Engineering and Technology, Peshawar, Pakistan.}
\address[2]{Department of Computer Science, Munsters Technological University, Cork, Ireland.}

\cortext[1]{Corresponding author.}
\fntext[1]{These authors contributed equally.}

\begin{abstract}
In recent years, social media has been widely explored as a potential source of communication and information in disasters and emergency situations. Several interesting works and case studies of disaster analytics exploring different aspects of natural disasters have been already conducted. Along with the great potential, disaster analytics comes with several challenges mainly due to the nature of social media content. In this paper, we explore one such challenge and propose a text classification framework to deal with Twitter noisy data. More specifically, we employed several transformers both individually and in combination, so as to differentiate between relevant and non-relevant Twitter posts, achieving the highest F1-score of 0.87.    

\end{abstract}

\maketitle

\section{Introduction}
\label{sec:intro}
Natural disasters, which are hazardous events and occur frequently in different parts of the world, can have devastating effects on society. Depending on the severity of the disaster, it may result in significant damage to the infrastructure and human lives. Rapid response to natural disasters may help in mitigating their adverse impact on society. In disasters and emergency situations, access to relevant and timely information is key to a rapid and effective response. However, the literature reports several situations where access to relevant and timely information may not be possible due to several factors \cite{ahmad2019social}.  

In recent years, social media outlets, such as Twitter, Facebook, and Instagram, have been explored as a source of communication and information dissemination in emergency situations \cite{said2019natural}. The literature already reports the feasibility and effectiveness of social media for a diversified list of tasks in disaster analytics. For instance, Ahmad et al. \cite{ahmad2017jord} explored social media outlets as a source of information collection and dissemination during natural disasters by proposing a system that is able to collect and analyze disaster-related multimedia content from social media. Similarly, social media content has also been utilized for disaster severity and damage assessment \cite{alam2017image4act,alam2018crisismmd}.

Despite being very effective in disaster analytics, social media data also come with several limitations. For instance, social media content contains a lot of noise and irrelevant information. This paper targets one of such challenges by proposing several solutions for the Relevance Classification of Twitter Posts (RCTP), sub-task introduced in DisasterMM challenge of MediaEval 2022 \cite{andreadis2022disastermm}. The task aims at automatically analyzing and classifying flood-related tweets into relevant and non-relevant tweets. 

\section{Related Work}
\label{sec:work}
 Disaster analysis in social media content has been one of the active topics of research in the domain over the last few years \cite{said2019natural}. During this time, different aspects and applications of disaster analytics in social media content have been explored \cite{ofli2020using}. Some key applications include communication/information dissemination, damage assessment, response management, sentiment analysis, and identification of the needs of affected individuals. The literature already reports several interesting works on these applications. For instance, Nguyen et al. \cite{nguyen2017damage} utilized social media content for damage assessment by analyzing disaster-related visual media posts. Ahmad et al. \cite{ahmad2019automatic} analyzed social media imagery for monitoring road conditions after floods. Moreover, a vast majority of the literature demonstrates how social media outlets can be used as means of communication in disasters and emergency situations \cite{palen2018social,ahmad2019social}. 

In the literature, different types of disasters including natural disasters, such as earthquakes, landslides, droughts, wildfires, and floods, as well as man-made disasters, such as accidents, have been explored \cite{ahmad2019social,ahmad2018comparative}. However, the majority of the works have targeted floods, being one of the most common natural disasters. The literature reports several interesting works on flood analysis in social media content for different tasks. For instance, Ahmad et al. \cite{ahmad2019automatic} proposed a late fusion-based framework for the automatic detection of passable roads after a flood. For this purpose, several deep learning models are trained on flood-related images from social media. Alam et al. \cite{alam2017image4act}, on the other hand, employed social media imagery for post floods damage severity assessment. 

Flood detection and analysis in social content have also been a part of the MediaEval benchmark initiative as a shared task for several years. Each time a separate aspect of flood analysis has been explored. For instance, in MediaEval 2017 the task aimed at the retrieval of flood-related images from social media. The task mainly involved analyzing the water level in different areas to differentiate between floods and regular water reservoirs, such as lakes \cite{bischke2017multimedia}. In MediaEval 2018, the task was slightly modified by asking the participants to propose multi-modal classification frameworks for flood-related multimedia content \cite{benjamin2018multimedia}. In MediaEval 2019 and 2020, the tasks aimed at analyzing flood severity and flood events recognition in social media posts.

\section{Approach}
\label{sec:approach}
\begin{figure*}[!h]
\centering
\includegraphics[width=0.6\textwidth]{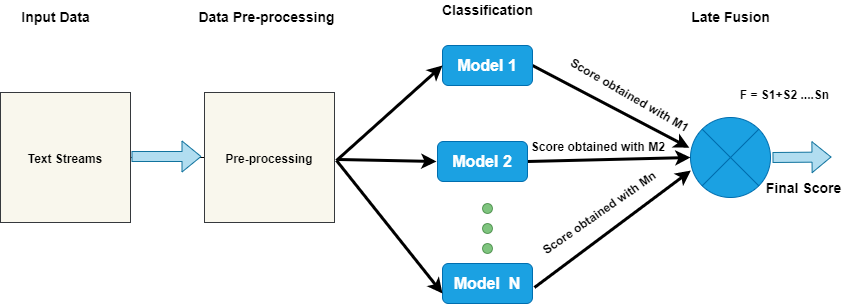}
\caption{Block diagram of the proposed approach.} 
	\label{fig:methodology}
\end{figure*}

Figure \ref{fig:methodology} provides the block diagram of the proposed framework for the RCTP task. The framework is composed of three main components namely (i) Pre-processing, (ii) Training and Classification, and (iii) Fusion. In the first step, different pre-processing techniques are employed to clean the dataset. Three different transformers are then trained on the data to obtain classification scores. In the final step, the classification scores of the individual models are combined in a late fusion scheme. The details of these steps are provided below. 

\subsection{Pre-processing}
In the pre-processing step, we employed different techniques for cleaning the dataset. More specifically, we removed unnecessary information, such as user names, URLs, emojis, punctuation marks, stop words, etc. Besides this, we also performed the necessary pre-possessing tasks that are required to transform the raw text into a form that is suitable for the transformers. To achieve this, we used the TF.text library\footnote{https://www.tensorflow.org/text/guide/bert\_preprocessing\_guide\#text\_preprocessing\_with\_tftext\#}. 

\subsection{Classification via Transformers}
After cleaning and pre-processing the data, we trained three different models, namely BERT \cite{devlin2018bert}, RoBERTa \cite{liu2019roberta}, and XLNet \cite{yang2019xlnet}. The selection of these models for the task is motivated by their proven performance on similar tasks \cite{ahmad2022social}. A brief overview of these models is provided below.

\begin{itemize}
    \item \textbf{BERT}: Bidirectional Encoder Representations from Transformers (BERT) is one of the state-of-the-art NLP algorithms for text processing. The model is pre-trained on a large collection of unlabeled text and can be fine-tuned for different text-analysis applications. The key attributes of the model include its bi-directional nature, pre-training with Masked Language Modeling (MLM), and Next Structure Prediction (NSP) objectives. 
    In the experiments with BERT, we used the Adam optimizer with a learning rate of 0.001 and a batch size of 8 for 3 epochs. 
     \item \textbf{RoBERTa}: Robustly Optimized BERT is a modified version of the BERT model with an improved training mechanism. More specifically, in RoBERTa the NSP capabilities are removed. Moreover, dynamic masking is introduced. In addition, a larger batch size and a  larger amount of training data were used in the training process. In this work, we used a learning rate of 0.001,  batch size of 20, and 10 epochs during the fine-tuning of the model for the desired task. 
     
      \item \textbf{XLNet}: XLNet is another state-of-the-art NLP algorithm. Similar to BERT, XLNet is also a bidirectional transformer and uses an improved training approach. In contrast to BERT and traditional NLP algorithms, XLNet relies on Permutation Language Modeling (PLM) by predicting all the tokens in random order. This allows XLNet to handle dependencies and bidirectional relationships in a better way. In this work, we used a learning rate of 0.002, a batch size of 32, and 4 epochs during the fine-tuning of the model for the desired task. 
\end{itemize}
We obtained the results in the form of posterior probabilities from these models, which are then used in the fusion scheme to obtain the final predicted labels. The fusion method used in this work is described in the next section.
\subsection{Fusion}
Our fusion method is based on late fusion, where we combined the classification scores obtained with the individual models for the final classification decision as shown in Equ. \ref{eqn:fusion}. In the equation, $S_{final}$ represents the final classification score while $s_{n}$ is the score obtained with the nth model. We note that in the current implementation, we used a simple fusion method by treating all the models equally (i.e., simple aggregation of the individual scores). 
\begin{equation}
\label{eqn:fusion}
S_{final}=S_{1}+S_{2}+s_{3}+....+S_{n}
\end{equation}

\section{Results and Analysis}
\label{sec:results}
Table \ref{tab:RCTP_dev} provides the experimental results of the proposed solutions on the development set. As can be been in the table, overall better results are obtained with the BERT model, and surprisingly, a lower F1-score is observed for RoBERTa. In the future, we will further investigate the potential causes of the lower performance of RoBERTa by exploring different implementations and hyper-parameter settings for it. As far as the performance of the fusion methods is concerned, overall better results are obtained with the pair of XLNet and BERT. One of the potential reasons for the lower performance of the fusion of all the models is the less accurate prediction of RoBERTa, as also evident from the performance of the individual models.

\begin{table}[!ht]
\caption{Experimental results of the proposed solutions on the development set.} 
\label{tab:RCTP_dev}
\begin{tabular}{cc}
\toprule
\textbf{Method} & \textbf{F1-Score} \\ 
\midrule
BERT      &     0.94      \\ 
RoBERTa    &     0.78     \\ 
XLNet   &   0.93      \\ 
Fusion 1 (RoBERTa, BERT, XLNet)  &  0.75     \\ 
Fusion 2 (BERT, XLNet)  &    0.93   \\
Fusion 3 (RoBERTa, XLNet)  &    0.92   \\ 
 \bottomrule
\end{tabular}
\end{table}
Table \ref{tab:RCTP_test} provides the official results of the proposed solutions on the test set. In total, three different runs were submitted. The first run is based on the fusion of all three models used in this work. The remaining two runs are based on the fusion of the models in pairs of two. In run 2, BERT and XLNet are combined while in run 3 RoBERTa and XLNet are jointly used. As can be seen in the table, better results are obtained for the fusion of the models in pairs of two where the best performing pair of two models obtained an improvement of 20\% over the fusion of all three models.

\begin{table}[!ht]
\caption{Experimental results of the proposed solutions on the test set.} 
\label{tab:RCTP_test}
\begin{tabular}{cccc}
   \toprule
\textbf{Run} & \textbf{Precision} & \textbf{Recall} & \textbf{F1-Score} \\
\midrule
 1 (Fusion of BERT, RoBERTa, XLNet)& 0.6738 & 0.5431 & 0.6014 \\ 
 2 (Fusion of BERT and XLNet) & 0.8044 & 0.6948 & 0.7456 \\ 
 3 (Fusion of RoBERTa and XLNet) & 0.8977 & 0.8598 & 0.8784 \\ 
  \bottomrule
\end{tabular}
\end{table}

\section{Conclusions}
In this paper, we presented our solutions for the RCTP subtask of DisasterMM challenge posted in MediaEval 2022. We proposed a late fusion framework incorporating several state-of-the-art transformers for the task. In the current implementation, all the models are treated equally by assigning them equal weights (i.e., 1). In the future, we aim to employ merit-based fusion methods to further improve the final classification score.
\def\bibfont{\small} 
\bibliography{references} 

\end{document}